\documentclass[a4paper]{llncs}
\usepackage{graphicx}
\usepackage{amsmath}
\usepackage{url}
 \usepackage[super]{nth}
 \usepackage{multirow}
\usepackage{tabulary}
\usepackage{float}

\title{Predicting Branch Visits and Credit Card Up-selling using Temporal Banking Data}
\author{Sandra Mitrovi\'c$^1$, Gaurav Singh$^2$}
\institute{
$^1$ KU Leuven (sandra.mitrovic@kuleuven.be) \\
$^2$ University College London (gaurav.singh.15@ucl.ac.uk)
}

\newcolumntype{K}[1]{>{\centering\arraybackslash}p{#1}}

\begin{document}
\maketitle
\begin{abstract}
There is an abundance of temporal and non-temporal data in banking (and other industries), but such temporal activity data can not be used directly with classical machine learning models. In this work, we perform extensive feature extraction from the temporal user activity data in an attempt to predict user visits to different branches and credit card up-selling utilizing user information and the corresponding activity data, as part of \emph{ECML/PKDD Discovery Challenge 2016 on Bank Card Usage Analysis}. Our solution ranked \nth{4} for \emph{Task 1} and achieved an AUC of \textbf{$0.7056$} for \emph{Task 2} on public leaderboard.
\end{abstract}

\keywords{Temporal Data, Feature Extraction, ECML Discovery Challenge}
\section{Introduction}
The number of tasks in which predictive analytics is used to support both business-as-usual and (strategic) decision making in different industries is increasing continuously.
In many service-oriented industries including banking, workload handling and customer satisfaction could be improved if company was able to predict customer behavior. For example, in banking, accurately predicting how many customers in certain period will visit a particular branch is very helpful to optimize workforce and logistics accordingly. On the other side, cost reduction and revenue management can also be improved knowing future behavior of a customer. Another banking example: predicting which customers will cross-sell/up-sell purchasing credit card, would allow marketing experts to use their resources more wisely when making decisions on future credit card campaigns.

The experiments done in this work are related to the two tasks of ECML/ PKDD Discovery Challenge 2016 on Bank Card Usage Analysis\footnote{More information on tasks, dataset and evaluation methods can be obtained at \url{https://dms.sztaki.hu/ecml-pkkd-2016/#/app/home/}}, which closely correspond to the above mentioned user behavior scenarios. More precisely:
\begin{itemize}
    \item \textbf{Task 1}: Predict the five most visited branches per user along with the corresponding number of visits  
    \item \textbf{Task 2}: Predict the future credit card buyers based on past activity
\end{itemize}

Since the majority of data provided for this competition consisted of time series data about users' activities, the most important challenge in this work has been to handle such time series data and use it for proposed regression and classification tasks. Rather than using sophisticated methods dedicated exclusively for time series analysis, as for example, AutoRegressive Moving Average (ARMA) or AutoRegressive Integrated Moving Average (ARIMA), our goal was to identify such features that would perform well with classical predictive models, such as Adaptive Boosting, Random Forest etc. Therefore, this work was focused on investigating how to extract information from temporal data in the form of fixed-length feature vectors, which both summarize and preserve discernible information, and which can subsequently be used with known machine learning models to solve regression and classification tasks. We should mention at this point that there have been some works on input representation learning from time series data using deep learning in the recent past \cite{zheng2014time}, but no general framework in the area of feature extraction from discrete temporal data has been developed to the best of our knowledge. 
%
%The dataset consists of three different segments: (1) Information about users (consisting of mostly binary variables with exception of geolocation and three categorical variables) (2) Time series data about users' activities (again consisting of geolocation and mostly categorical variables) and 3) Bank geolocation data.
%

The rest of the paper is organized as follows: in Section \ref{sa} we explain our methodology, in Section \ref{res} we present and discuss obtained experimental results and  in Section \ref{conc} we conclude and discuss possible improvements.
\section{Proposed Method}
\label{sa}
We explored two different directions of experimentation: 1) with extracting new features and 2) with different regression (for Task 1) and classification (for Task 2) models. In some of the experiments, we also used (unsupervised) clustering techniques, as it would be explained later in Sections \ref{feat} and \ref{res}. 
\subsection{Data Pre-processing}
\label{dp}
User info and activity data contained some missing values denoted with "-". Additionally, there were users without any activity (8,646 out of total 191,238 train users and 44,916 out of 191,237 test users, that is 4.5\% for train and 23.5\% for test set). We handled these by replacing missing values with brand new categories for categorical variables, while for other variables we simply replaced missing values with zeros. On top of this, in test set, "AGE\_CAT" feature contained some missing values but since these were few (only 638 out of 191,237), we replaced these with the most frequent category (which happened to be "b" (36-65) with 54.82\% with 104,832 users compared to "a" ($\le35$) with 58,834 and "c" ($\ge 65$) with 26,933 users). We performed label encoding and one-hot encoding for all categorical features. %We experimented with scaling for generalized models and logarithm for power-law distributed features, but without success. 
Since no branch related activities were available in the test set, we completely omitted this data during model construction phase. For the same reason, we were not considering data referring to credit card usage and wealth of customer for second half of 2014. %In some cases, due to power law distribution of features, we applied logarithmic transformation, as it will be mentioned later in \ref{res}. %We also investigated how the results change with and without scaling.
\subsection{Feature Extraction}
\label{feat}
Our idea was to start simple and gradually generate and employ additional informative features, %After initial attempts using only user info data, we started exploiting activity data as well.
% Our main focus was on extracting all possible informative features from activity data
mainly from activity data, %from which we extracted a number of informative features 
with intention to retain particularities of distinct activities. Following list provides feature extraction sequence and feature sets (FS) for our cross-validation experiments (each step, except the first one, comprises previous step features enriched with new one(s)):
\begin{itemize}
\item FS1: Averaged values of activity features per user\\
    (note that we already transformed categorical features, see Section \ref{dp})
\item FS2: Minimal distance between user geolocation and branches geolocations
\item FS3: Set of features (various counters) per user:
    \begin{itemize}
        \item Number of POS activities
        \textbullet~Number of web shop activities
        \textbullet~Number of credit card usages
        \textbullet~Number of debit card usages
        \textbullet~Number of different amount categories during activities
        \textbullet~Maximal amount category used during activities
        \textbullet~Number of days since the last activity 
        \textbullet~Number of different activity locations
        \textbullet~Number of different activity time slots per user
        \textbullet~Number of different marketing categories 
        \textbullet~Total number of user activities
        \textbullet~Most frequent activity time slot 
        \textbullet~Most frequent activity location category 
        %\textbullet~Indicator if the activity happened during January 
        %\textbullet~Indicator if the activity happened during February-June
        \textbullet~Number of months user possesses credit card
    	\textbullet~Number of months user is categorized as "wealthy"
    \end{itemize}
\item FS4: Three features based on inter-activities time (measured in days):
    \begin{itemize}
        \item Mean inter-activity time per user 
        \textbullet~Standard deviation of inter-activity time per user
        \textbullet~User $u$ activity clumpiness \cite{Clumpiness} $C_u$, defined as: 
        $ C_u = 1+\frac{\sum_{i=1}^{n^u+1}log(x^u_i) x^u_i}{log(N+1)}$,
        where $x^u_i$ is $i^{th}$ occurrence of activity for user $u$, $n^u$ is number of activities of user $u$ and N is potential number of activities (since we consider only days and first half of the year, in our case N=181)  
    \end{itemize}
\item FS5: Two features: %  act_dist + ratio
%\begin{itemize}
    \textbullet~Average Euclidean distance between user geolocation and user activity geolocation
    \textbullet~Average ratio of user location-activity distance and total number of user activities (zero, if user had no activities) % ask G 
%\end{itemize}
\item FS6: Two trend features per each of "AMT\_CAT" and "MC\_CAT":
%\begin{itemize}
    \textbullet~Ratio of positive changes of variable in user activity sequence
    \textbullet~Ratio of negative changes of variable in user activity sequence \\
Despite the absence of exact amount, "AMT\_CAT" is ordinal variable taking values "a"-low, "b"-medium, "c"-high, so we could observe positive/negative trends; for "MC\_CAT", we assumed that categories "a"-"j" also represent some kind of ordinal data (e.g. increasing solvency of the user)
 \item FS7: Minimal Euclidean distance between mean user activity geolocation and branches geolocations
 \item FS8: Cluster assignment based on $k$-means clustering of users geolocations
 \item FS9: Distance between user geolocation and geolocation of the branch for which we try to make visits prediction
 \item FS10: Distance between mean user activity geolocation and geolocation of the branch for which we try to make visits prediction
%
%none	 - Y normalization
%none	currently best submission
\end{itemize}

\subsection{Models}
\label{meth}
For model construction for Task 1, we experimented with several regressors available in Python Scikit library \cite{sklearn_api}: Random Forest Regressor (RFR \cite{breiman2001random}), Adaptive Boosting Regressor (ABR) and Gradient Boosting Regressor (GBR). We trained 323 regressors (1 per each branch), selecting as final result 5 most visited branches. These 323 models were processed in parallel using multiprocessing\footnote{Code can be found at \url{https://github.com/SandraMNE/ECMLChallenge2016}}. We also experimented with Neural Nets (NN) using Theano-based Lasagne \cite{lasagne} library. For Task 2, we used equivalent classification models (RFC \cite{breiman2001random}, ABC \cite{freund1995desicion}, GBC \cite{friedman2001greedy}), Logistic Regression and Neural Nets.   
\section{Experimental Results}
\label{res}
Apart from mentioned supervised techniques, in our initial attempts we tried applying $k$-means clustering on user info data for Task 1, hoping that similarities between users could prove to be significant. We also experimented with different number of clusters using BIRCH \cite{BIRCH} clustering algorithm and tried to combine clustering results obtained using only user geolocation features with clustering results obtained using all available user information. Unfortunately, and despite trying different number of clusters, in both cases results were unsatisfactory.

We also experimented with weighted averaging of activities for a given user whereby the weights were based on the logarithm of the recency of an activity. To compute the recency of a given activity we computed the number of days passed from a reference date. We expected that giving higher weights to more recent activities would lead to better results, but the final results showed that a simple unweighted averaging performed better. 

The two-fold cross validation results obtained per Task 1 and Task 2 using different methods and feature sets can be seen in Table \ref{table:resT1} and Table \ref{table:resT2}, respectively. We omitted Neural Network results for Task 1 and $k$-means results for both Task 1 and Task 2, since they performed drastically worse than the other methods, despite our efforts to tune parameters. The best result for Task 1 (both for cross validation and on public leaderboard) was achieved using GBR with all generated features and normalizing target variable. We also tried stacking a Ridge regression on the output of GBR to utilize the dependence between branch visits, but, contrary to our expectations, the results did not indicate the presence of such dependence. 

For Task 2, displayed results for Neural Network refer to one hidden layer with tangent hyperbolic non-linearity. We experimented with dropout for regularization, but without success. We did not consider FS9 and FS10 interesting for Task 2. The best cross validation result for Task 2 was obtained for LR for FS7, followed closely by LR with FS8. Our best public board submission was obtained by performing simple unweighted ensemble of the following four submissions: 1) LR with FS8 with feature scaling, 2) GBC with FS1, 3) GBC with FS8 but without any geolocation coordinates, 4) GBC with FS8 but without C201* and W201* columns. 
However, this approach scored only 0.0001 better than GBC with FS1 when applying logarithmic transformation on input features. It is worth mentioning that we achieved better performance by considering as positive those cases when credit card was bought in 2015, instead of just restricting them to second half of 2014.

\begin{table} [ht]
\centering
\caption{Task 1 cross validation results for different feature sets (FS) and models. Evaluation measures are the same as the ones used for public leaderboard evaluation. Tuned number of base estimators was 100 for all methods, except with ABR where 10 base estimators performed the best.}
\begin{tabular}{|c||c|c|c|} 
 \hline
\multirow{2}{*}{Feature set} & 
\multicolumn{3}{c|}{Task-1 (Cosine-based)} \\
\cline{2-4}
 & Gradient Boosting & Adaptive Boosting & Random Forest \\ %[12pt]\\ 
\hline \hline
FS1 & 0.59977 & 0.43163 & 0.61358 \\ %& \\[0.1ex] 
\hline
FS2 & 0.60421 & 0.41698 & 0.61502 \\ %& \\[0.1ex]
\hline
FS3 & 0.60532 & 0.43188 & 0.61419 \\ %& \\[0.1ex]
\hline
FS4 & 0.60347 & 0.42777 & 0.61389 \\ %& \\[0.1ex]
\hline
FS5 & 0.59681 & 0.42099 & 0.60740 \\ %& \\[0.1ex]
\hline
FS6 & 0.59526 & 0.40929 & 0.60594 \\ %& \\[0.1ex]
\hline
FS7 & 0.59545 & 0.41822 & 0.61132 \\ %& \\[0.1ex]
\hline
FS8 & 0.61190 & 0.46184 & 0.61325 \\ %& \\[0.1ex]
\hline
FS9 & 0.62478 & 0.48508 & 0.61718 \\ %& \\[0.1ex]
\hline
FS10 & 0.62482 & 0.49016 & 0.61879 \\ %& \\[0.1ex]
\hline
FS10 + normal. & \textbf{0.65167} & 0.50470 & 0.63154 \\ %& \\[0.1ex]
\hline
\end{tabular}
\label{table:resT1}
\end{table}
\begin{table} [ht]
\centering
\caption{Task 2 cross validation results for different feature sets (FS) and models. Evaluation measures are the same as the ones used for public leaderboard evaluation, and the number of base estimators was obtained as 100 after tuning for optimum results.}
\begin{tabular}{|K{2cm}||K{1.5cm}|K{1.5cm}|K{1.5cm}|K{1.5cm}|K{1.5cm}|} 
 \hline
\multirow{2}{*}{Feature set} &  
\multicolumn{5}{c|}{Task-2 (AUC)} \\
\cline{2-6}
 & LR & ABC & RFC & NN & GBC \\ %[12pt]\\ 
\hline \hline
FS1 & 0.69602 & 0.70858 & 0.65534 & 0.63450 & 0.71404 \\ %& \\[0.1ex] 
\hline
FS2 & 0.70780 & 0.69388 & 0.64498 & 0.61947 & 0.70067 \\ %& \\[0.1ex]
\hline
FS3 & 0.70798 & 0.70022 & 0.65116 & 0.63219 & 0.70828 \\ %& \\[0.1ex]
\hline
FS4 & 0.70771 & 0.70209 & 0.65229 & 0.63538 & 0.70781 \\ %& \\[0.1ex]
\hline
FS5 & 0.70815 & 0.69579 & 0.64207 & 0.60596 & 0.70403 \\%& \\[0.1ex]
\hline
FS6 & 0.70737 & 0.70312 & 0.64397 & 0.60507 & 0.70913 \\%& \\[0.1ex]
\hline
FS7 & \textbf{0.72871} & 0.70110 & 0.64361 & 0.61467 & 0.71238 \\%& \\[0.1ex]
\hline
FS8 & 0.72866 & 0.70672 & 0.65033 & 0.61193 & 0.71381 \\%& \\[0.1ex]
\hline
\end{tabular}
\label{table:resT2}
\end{table}

We observed that our cross validation and public leaderboard results did not display complete coherence. For example, for Task 1 and feature set FS10 with applied normalization, GBR result on cross validation set was 0.6517 while the result obtained on public leaderboard was 0.6561. On the other hand, for Task 2, there were even greater disparities between cross validation and submission scores. For example, for GBC with FS1 we achieved only 0.7055 on public leaderboard, despite our cross validation result of 0.71404.
Due to these fluctuations, even though the addition of some features decreased the performance by a small amount in our cross validation, we did not remove those features in public submissions.  
%

%
%
%\begin{table} [ht!]
%\centering
%\caption{Task 1 \& Task 2 cross validation (CV) and public leaderboard (PL) results for several feature sets (FS). For Task 1, GBR results are displayed. For Task 2...}
%\begin{tabular}{|c||c|c||c|c|} 
% \hline
%\multirow{2}{*}{Feature set} & %\multicolumn{2}{|c||}{Task-1}  & %\multicolumn{2}{|c|}{Task-2} \\
%\cline{2-5}
% & CV & PL & CV & PL \\ %[12pt]\\ 
%\hline \hline
%FS1 & 0.59977 &  &  & \\ %& \\[0.1ex] 
%\hline
%FS2 & 0.60421 &  &  &  \\ %& \\[0.1ex]
%\hline
%FS3 & 0.60532 &  &  &  \\ %& \\[0.1ex]
%\hline
%FS4 & 0.60347 &  &  &   \\ %& \\[0.1ex]
%\hline
%FS5 & 0.59681 &  &  &   \\%& \\[0.1ex]
%\hline
%FS6 & 0.59526 &  &  &   \\%& \\[0.1ex]
%\hline
%FS7 & 0.59545 &  &  &   \\%& \\[0.1ex]
%\hline
%FS8 & 0.61190 &  &  &   \\%& \\[0.1ex]
%\hline
%FS9 & 0.62478 &  &  &   \\%& \\[0.1ex]
%\hline
%FS10 & 0.62482 &  &  &   \\%& \\[0.1ex]
%\hline
%FS10 + normal. & 0.6517 & 0.6561 &  &   \\%& \\[0.1ex]
%\hline
%\end{tabular}
%\label{table:res}
%\end{table}
%
\section{Conclusion}
\label{conc}
In this work, we extract different components of the temporal data to prepare fixed-length feature vectors that can be utilized with well-known machine learning models.
One limitation of our approach was to use in submissions all generated features, instead of performing feature selection for recovering the optimal set of features for each task. We believe that feature selection could have led to further improvements in our public leaderboard results, but due to time constraints, we focused our attention on feature extraction from the temporal data. Accounting for seasonality effects could probably as well increase the performance.
Public leaderboard results suggest that this level of complexity suffices for scoring high in Task 1. For Task 2, our cross validation results indicate an even better performance than obtained on public leaderboard, especially as current public leaderboard evaluation is based on only 30\% of test dataset. 

\bibliographystyle{splncs03}
\bibliography{main}

\begin{thebibliography}{1}
\providecommand{\url}[1]{\texttt{#1}}
\providecommand{\urlprefix}{URL }

\bibitem{breiman2001random}
Breiman, L.: Random forests. Machine learning  45(1),  5--32 (2001)

\bibitem{sklearn_api}
Buitinck, L., Louppe, G., Blondel, M., Pedregosa, F., Mueller, A., Grisel, O.,
  Niculae, V., Prettenhofer, P., Gramfort, A., Grobler, J., Layton, R.,
  VanderPlas, J., Joly, A., Holt, B., Varoquaux, G.: {API} design for machine
  learning software: experiences from the scikit-learn project. In: ECML PKDD
  Workshop: Languages for Data Mining and Machine Learning. pp. 108--122 (2013)

\bibitem{lasagne}
Dieleman, S., Schlüter, J., Raffel, C., Olson, E., Sønderby, S.K., Nouri, D.,
  Maturana, D., Thoma, M., Battenberg, E., Kelly, J., Fauw, J.D., Heilman, M.,
  diogo149, McFee, B., Weideman, H., takacsg84, peterderivaz, Jon, instagibbs,
  Rasul, D.K., CongLiu, Britefury, Degrave, J.: Lasagne: First release. (Aug
  2015)

\bibitem{freund1995desicion}
Freund, Y., Schapire, R.E.: A desicion-theoretic generalization of on-line
  learning and an application to boosting. In: European conference on
  computational learning theory. pp. 23--37. Springer (1995)

\bibitem{friedman2001greedy}
Friedman, J.H.: Greedy function approximation: a gradient boosting machine.
  Annals of statistics pp. 1189--1232 (2001)

\bibitem{BIRCH}
Zhang, T., Ramakrishnan, R., Livny, M.: Birch: an efficient data clustering
  method for very large databases. In: ACM Sigmod Record. vol.~25, pp.
  103--114. ACM (1996)

\bibitem{Clumpiness}
Zhang, Y., Bradlow, E.T., Small, D.S.: Predicting customer value using
  clumpiness: From rfm to rfmc. Marketing Science  34(2),  195--208 (2015)

\bibitem{zheng2014time}
Zheng, Y., Liu, Q., Chen, E., Ge, Y., Zhao, J.L.: Time series classification
  using multi-channels deep convolutional neural networks. In: International
  Conference on Web-Age Information Management. pp. 298--310. Springer (2014)

\end{thebibliography}
\end{document}